\documentclass[12pt]{article}
\usepackage{arxiv}
\usepackage{graphicx} 
\usepackage{tikz}
\usepackage{url}
\usepackage{booktabs}
\usepackage{subcaption}
\usepackage{adjustbox}
\usepackage{multirow,multicol}

\newcommand{\bt}{\mathbf{t}}

\newcommand{\bY}{\mathbf{Y}}
\newcommand{\bW}{\mathbf{W}}
\newcommand{\bX}{\mathbf{X}}

\usepackage{amsmath,amsfonts}

\title{NNTile: a machine learning framework capable of training extremely large GPT language models on a single node}

\newcommand{\shorttitle}{NNTile: a machine learning framework}

\author{Aleksandr Mikhalev\thanks{Corresponding author}\\
Skoltech, Moscow, Russia \\
\url{al.mikhalev@skoltech.ru}\\
\And 
Aleksandr Katrutsa\\
Skoltech, Moscow, Russia\\
AIRI, Moscow, Russia\\
\url{amkatrutsa@gmail.com}\\
\And
Konstantin Sozykin\\
Skoltech, Moscow, Russia\\
\url{ko.sozykin@skoltech.ru}\\
\And
Ivan Oseledets\\
AIRI, Moscow, Russia\\
Skoltech, Moscow, Russia\\
\url{oseledets@airi.net}
}

\begin{document}
\maketitle

\begin{abstract}
This study presents an NNTile framework for training large deep neural networks in heterogeneous clusters. The NNTile is based on a StarPU library, which implements task-based parallelism and schedules all provided tasks onto all available processing units (CPUs and GPUs). 
It means that a particular operation, necessary to train a large neural network, can be performed on any of the CPU cores or GPU devices, depending on automatic scheduling decisions.
Such an approach shifts the burden of deciding where to compute and when to communicate from a human being to an automatic decision maker, whether a simple greedy heuristic or a complex AI-based software.
The performance of the presented tool for training large language models is demonstrated in extensive numerical experiments.
\end{abstract}

\section{Introduction}

Over the last few years, the global market of artificial intelligence (AI) has grown significantly. 
The trend is still going on: judging by \url{MarketsAndMarkets.com}, global AI market value was estimated at \$150.2 billion in 2022~\cite{mam-2023} with a projected value of \$1345.2~\cite{mam-2023} billion by 2030 and a Compound Annual Growth Rate (CAGR) of 35.8\%.
The main reason for such a rise is a ``Generative AI'', backed by transformers neural network architecture~\cite{vsp-2017-transformers}.
Corresponding to~\cite{kmh-2020-scaling}, one just puts more parameters into a transformer, feeds it more data, and spends more computing power to improve outcomes.
The optimal amount of data and compute for a model of a given size is studied in~\cite{hbm-2022-chinchilla}.

Just like any other neural network architecture nowadays, transformers are pretrained on thousands of GPUs, while CPUs are heavily underutilized, and CPU RAM is not taken into account at all.
Here's a simple example: a single DGX-A100 workstation contains 8 Nvidia A100 GPUs with 80GB VRAM and 2 TB of CPU RAM.
Popular solutions for distributed training of neural networks, e.g., DeepSpeed~\cite{rrr-2020-deepspeed,aminabadi2022deepspeed} or Megatron-LM~\cite{spp-2019-megatron,spn-2022-megatron,kcl-2023-megatron} or MegaScale~\cite{jlz-2024-megascale} or PyTorch~\cite{paszke2019pytorch} FSDP~\cite{zgv-2023-fsdp}, are limited by an aggregated 640GB of VRAM in such a case because offloading to CPU RAM causes synchronizations and, as a result, a significant loss of a training throughput.
A comprehensive survey on the main approaches to distributed training of large models is presented in~\cite{gusak2022survey}.
However, if training requires more than 640 GBs of aggregated memory, then either training does not run at all or overall performance drops drastically.
In the case of a cluster of GPUs, if aggregated VRAM is not enough, then there is only one solution -- to get more GPUs to avoid the performance drop.
Without a doubt, getting more GPUs for the pretraining phase to digest terabytes of data is the way to go, and at the first glance, there is no trouble at all.
However, such a memory wall defines a minimal number of GPUs for all other phases related to training: searching for training hyperparameters and fine-tuning for downstream tasks, especially if a model supports very long sequences of data.
Acquiring thousands of GPUs just to check training loss behaviour at another value of a learning rate might postpone a model release date heavily.

Although pretraining a transformer model from a scratch or a checkpoint does not suffer the aforementioned memory wall, average model floating point operations (model FLOPs) per GPU barely reach 55\% of the theoretical peak.
For example, Megatron-LM reports training efficiency for a model with 530 billion parameters at 40.4\% on 2240 GPUs, at 38.8\% on 2800 GPUs, and at 36.2\% on 4480 GPUs in~\cite{spn-2022-megatron} and at 56\% on 280 GPUs in \cite{kcl-2023-megatron}.
Technical report~\cite{aaa-2023-falcon} on training a Falcon model with 40 billion parameters along with its model card on HuggingFace~\footnote{\url{https://huggingface.co/tiiuae/falcon-40b}} states that 2800 PFlops-days were digested on 384 Nvidia A100 GPUs through around two months, which translates into approximately 40\% efficiency.
Recent paper~\cite{jlz-2024-megascale} claims to get 54.3\% efficiency on training 530-billion parameter model on 11200 GPUs and 55.2\% efficiency on training 175-billion parameter model on 12288 GPUs, while Megatron-LM achieves only 48.2\% and 41.2\% respectively.

In this paper, we present a non-standard approach to train neural networks.
To the best of our knowledge, this is the first study of a task-based parallel programming paradigm applied to train transformer-based models.
Such a paradigm distributes computations among available resources dynamically and prefetches and offloads data asynchronously.
It is described in more details section~\ref{sec:task-based}.
Such a paradigm shift allows us to train custom GPT2 models with up to 50 billion parameters on a single server.
In the meantime, a popular FullyShardedDataParallel (FSDP) approach of PyTorch allows only training of models
up to 8 billion parameters while all other settings are equal.

The main contributions are listed below.
\begin{enumerate}
    \item The brief introduction to the task-based parallelism and StarPU~\cite{starpu} library is presented.
    \item The implementation details of the key elements of deep neural networks in a task-based paradigm are discussed.
    \item The first experimental results on using task-based parallelism for training the GPT2 model in the single node are shown.
\end{enumerate}



\section{Task-based parallelism}
\label{sec:task-based}
We consider a computation process as a sequence of operations and necessary data for it.
It is sometimes convenient to split huge operations into smaller tasks operating on pieces of data.
For example, it is natural to split linear algebra operations, like matrix multiplication or factorization, into their corresponding block versions.
Block versions split entire data into chunks, referred to as \emph{tiles}, and initial operations into a set of tasks, where each task has a limited data accesses to just few tiles.
These tasks are composed together into a directed acyclic graph (DAG), where nodes correspond to tasks and directed edges correspond to tiles.
The smaller the tiles are, the more tasks are present in the DAG.
As a consequence, more tasks can be executed in parallel, but arithmetic intensity along with performance of each task may drop down.
Once the DAG of tasks is formed, a scheduler comes into play and maps each tasks onto available computational resources dynamically during runtime.
The scheduler also transfers necessary tiles to and from designated devices asynchronously.
Further, we cover all the main ingredients of the task-based parallel programming paradigm: task submission, memory management and task scheduler.

\textbf{Task submission.}
There are two main approaches to manage the task submission process and compose the computational graph.
The first is Sequential Task Flow (STF), which inserts tasks into the graph one by one during runtime.
This approach emulates the sequential running of the tasks, while they are executed asynchronously due to readiness of the input tiles.
Such an emulation significantly simplifies the source code reading and modifying.
The second approach is called a Parametrized Task Graph (PTG), which describes the computational graph parametrically.
The PTG allows to instantiate a portion of a graph from its parametric description, in contrast to the STF approach.
In addition, dependencies between subtasks are known in advance and, therefore, collective MPI routines can be used.
Our work, described in the current paper, focuses on the STF approach and its implementation within the StarPU library~\cite{starpu}.

\textbf{Memory management.}
There are 2 ways to work with the data: offload it to a GPU from the main memory for every single task or hold it on a GPU while it is still needed.
The offloading approach proved to be easy to implement and useful in a case of standard dense linear algebra applications.
However, neural networks training workflow suffers from a much low arithmetic intensity.
It bottlenecks the offloading approach by a CPU-GPU communication bus.
Therefore, we use a memory manager that preserves tiles on a GPU if needed for further tasks.
Tiles can be asynchronously copied to another GPU if necessary.
The StarPU library supports such a memory management system, which we rely on in our numerical experiments.

\textbf{Schedulers.}
The next ingredient of the framework for a task-based parallelism is the scheduler, which decides which device runs the current task and how to minimize the total runtime based on data communication costs, task complexity, and device availability.
Simple yet effective greedy heuristics proved to fit standard dense linear algebra applications through the StarPU and Chameleon~\cite{agullo2012hybridization} software.
Such heuristics schedule tasks one by one only taking into account already scheduled tasks without looking into the forthcoming remaining graph.
The main reason for such heuristics to shine is within an incredibly high arithmetic intensity of the overall problem.
Although training neural networks is mostly memory bound, entire graph of tasks is made of many repeated training iteration.
Such an assumption allows to develop a so-called graph scheduling policy, that takes into account entire graph to make scheduling decisions.

\section{GPT2 model training with NNTile.}

To illustrate the efficiency of our approach, we train custom GPT2 models with an increasing number of parameters and compare the maximum feasible model size with PyTorch in FSDP mode.
We re-implement all the necessary operations to train the GPT2 model in the tile-based form within the \emph{NNTile} machine learning framework and list them below.
Due to page limitation, we describe only forward passes through layers of the GPT2 model, although operations for gradient propagation from the last layer to the first one are also implemented in NNTile.

\textbf{Custom GPT2 models configuration.}
Unfortunately, training standard pretrained GPT2 models of a small size, available at the Huggingface portal~\footnote{https://huggingface.co/openai-community/}, did not scale from a single GPU to 8 GPUs on a single node in a data parallel training regime.
Calculation of gradients of model parameters becomes a bottleneck in such a case.
An obvious solution is to support a reduction: gradients of the same matrix of parameters from different devices are accumulated.
However, enabling reduction with a help of StarPU data access mode STARPU\_REDUX made things even worse.
Greedy scheduling policies fail to get good scaling in such a scenario.
Therefore, it was decided to train custom GPT2 models with just few transformer blocks but with large embedding size to split entire data into tiles across embedding dimension.
Our numerical examples proved that standard StarPU scheduling policies are more or less good fit for such a \emph{tensor} parallelism.

\textbf{Embedding layer.}
This layer processes the input matrix of size $N_{s} \times N_b$, where $N_b$ denotes batch size and $N_s$ denotes the sequence length and equips every tokenized item from a sequence with an embedding vector. 
The tokenized item of the sequence corresponds to the map of a word from a sequence given by the selected tokenizer.
The resulting three-dimensional tensor is of the size $N_{e} \times N_{s} \times N_b$, where $N_e$ denotes the embedding dimension.
So, this layer has $N_v \times N_e$ parameters, and $N_v$ is the vocabulary size aligned with the tokenizer.

\textbf{Nonlinear activations.}
Most of non-linear activation functions used in deep neural networks are elementwise transformations of the input data.
Therefore, they perfectly fit the tile representation of the tensor used in our framework.
Low-level kernels, corresponding to activations, apply directly to every tile.




\textbf{Linear layer.}
This layer in the GPT2 model transforms the embedding dimension of the input data $\mathbf{X}$ from $N_e$ to $N_d$ via a linear transformation.
Formally, this transformation is done through a matrix multiplication and an addition of a bias over unchanged dimensions: $\bY = \bW\bX \mathop{+.} \mathbf{b}$, where $\bY$ is of size $N_d \times N_s \times N_b$, $\bW$ is of size $N_d \times N_e$ and $+.$ denotes the summation of the same data over unchanged dimension.
Tile-based implementation of this layer is also straightforward and based on the multiplications and additions of elementary tiles.



\textbf{Layer normalization layer.}
Normalization layers are designed to improve stability of the training process and avoid gradient vanishing/exploding phenomena.
Since our primary focus is the transformer-based models, we consider the Layer Normalization layer~\cite{ba2016layer}.
This type of normalization shows superior performance for transformer models over alternative normalization layers~\cite{xiong2020layer}.
The tile-based implementation of this layer consists of three steps: mean and variance of every tile accumulation, normalization of every tile and scaling with an addition of bias.

\textbf{SoftMax.}
SoftMax transformation converts the samples' embeddings to probabilities.
Formally, given embedding $\bt \in \mathbb{R}^C$, where $C$ is a number of classes, for example, the SoftMax estimates probabilities $\hat{p}_i$ as 
$
\hat{p}_i = \frac{e^{t_i}}{\sum_{k=1}^C e^{t_k}} = \frac{e^{t_i - t_{\max}}}{\sum_{k=1}^C e^{t_k - t_{\max}}},
$
where $t_{\max} = \max_i t_i$.
The latter equality is an implementation detail to avoid infinities while calculating sum of exponents.
Therefore, the computing of SoftMax for tensor decomposed in tiles consists of two subroutines.
The first one computes $t_{\max}$ and denominator per tile, and the second one aggregates them with entire values stored in tiles and computes the target probabilities estimate.
In the GPT2 model, SoftMax is used inside the Attention layer (see below) and in the cross-entropy loss function.

\textbf{Attention layer.}
This layer is crucial in constructing efficient embeddings for solving NLP tasks.
The attention layer with a head size $h$ contains three matrix multiplications to construct embeddings of queries $\mathbf{Q}$, keys $\mathbf{K}$ and values $\mathbf{V}$.
These embeddings are of size $h \times N_{s} \times N_{b} \times N_h$, where $N_{s}$ denotes the number of sequences, $N_b$ is the number of batches, and $N_h$ is the number of heads.
Then the SoftMax is applied: $\mathbf{B} = \mathbf{V}\left(\mathrm{SoftMax}\left(\frac{1}{\sqrt{h}}\mathbf{K}^\top \mathbf{Q}\right)\right)$ and the resulting embeddings $\mathbf{Y}$ are computed via linear projection $\mathbf{Y} = \mathbf{W}\mathbf{B}^\top$, where $\mathbf{W} \in \mathbb{R}^{N_e \times N_h \times h}$.
The resulting embedding tensor~$\mathbf{Y}$ is of the size $N_{e} \times N_{s} \times N_b$.
We implement these transformations based on tile splitting of input data and parameters.

\textbf{Cross-Entropy loss} is a multiclass generalization of the binary logistic loss, which measures the prediction quality in the training stage.
It is computed via following equation $L =  -\sum_{i=1}^{N_b}-\log\left( \frac{e^{x_{ic}}}{\sum_{j=1}^C e^{x_{ij}}} \right)$, where $c$ is a correct class label and $x_{ij}$ is proportional to the probability of the $i$-th sample to be assigned to the $j$-th class.
Computing this loss is also feasible for input embeddings split on tiles. 

\textbf{Optimizers.}
The standard optimizers in training deep neural networks are SGD with momentum~\cite{lin2020accelerated}, Adam~\cite{kingma2014adam}, and its modification called AdamW~\cite{loshchilov2017fixing}.
They minimize the loss function~$\mathcal{L}$. 
In particular, an update step for SGD is just a weighted sum of the input vector, gradient, and momentum term.
This update step perfectly fits a tile decomposition of the used tensors.
Although the update steps for the Adam and AdamW optimizers are more tricky and require twice as much memory compared to SGD with momentum, they are also consistent with the tile decomposition.

\textbf{NNTile distribution.}
The NNTile is an open source project.
It is under development right now, with very limited functionality.
The NNTile is available as the GitHub repository at \url{https://github.com/nntile/nntile}.

\section{Numerical experiments}
\begin{figure}
    \centering
    \includegraphics[width=0.7\linewidth]{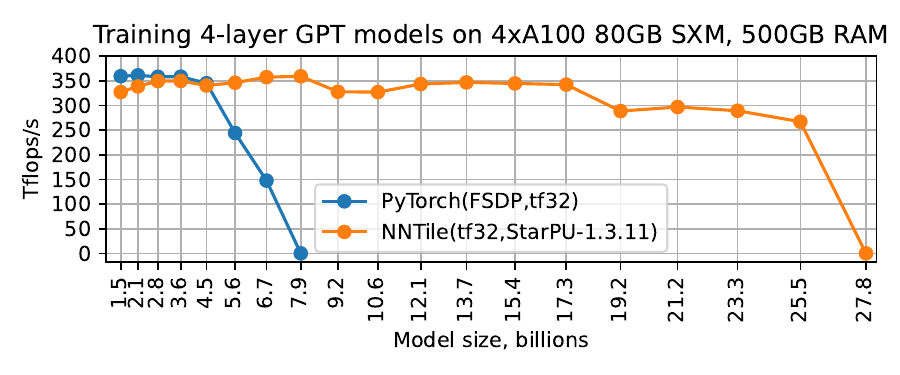}
    \caption{GPT2 model with 4 layers trained on the four A100 GPUs in tf32 format of floating point numbers. NNTile can train significantly larger models than PyTorch FSDP (25.5B vs. 6.7B). Zero Tflops/s indicates Out-Of-Memory error.}
    \label{fig:4layers_4gpus}
\end{figure}

\begin{figure}[!ht]
    \centering
    \includegraphics[width=0.7\linewidth]{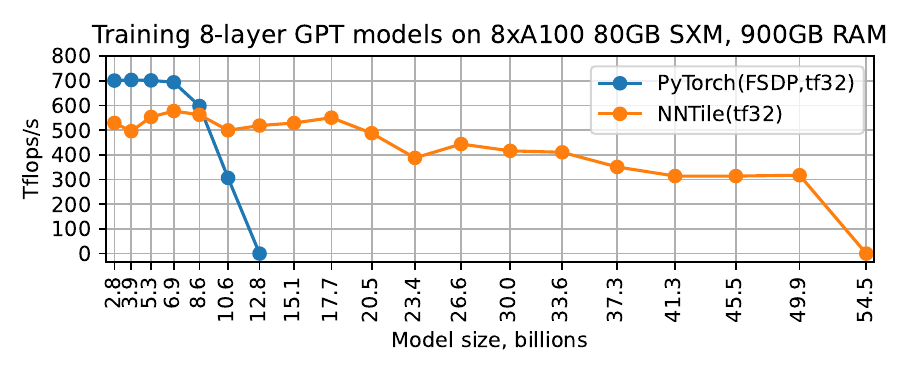}
    \caption{GPT2 model with 8 layers trained on eight A100 GPUs in tf32 format of floating point numbers. NNTile can train significantly larger models than PyTorch FSDP (49.9B vs. 10.6B). Zero Tflops/s indicates Out-Of-Memory error.}
    \label{fig:8layers_8gpus}
\end{figure}

In the numerical experiments, we compare the performance and the scalability of the GPT2 models with 4 and 8 layers of various embedding sizes trained with NNTile and PyTorch.
All the experiments were conducted with an NNTile version 1.0.0 (\url{https://github.com/nntile/nntile/tree/1.0.0}) on a single node with 8 A100 GPUs.
Figure~\ref{fig:4layers_4gpus} shows that NNTile performs on par with PyTorch FSDP, while being able to train much larger models on the same hardware.
This is due to a built-in automatic offloading of data from VRAM of GPUs to CPU RAM while CPU cores are idle.
Similar observation holds for GPT2 with 8 layers; see Figure~\ref{fig:8layers_8gpus}.
However, 8-layer model suffers from performance degradation due to far from optimal decisions of StarPU greedy scheduling policy \emph{dmdasd}.
Unfortunately, the more layers are in a neural network, the worse performance we get with the NNTile.

\section{Conclusion}
This paper introduces the task-based parallelism paradigm for training large language models in a heterogeneous environment and the NNTile machine learning framework built on top of the StarPU library.
The NNTile framework supports basic ingredients for neural network training.
We implement forward and backward passes for the most popular layers (e.g., linear layer, attention layer, layer normalization layer, etc.), optimization methods based on the stochastic gradient estimate (e.g., Adam, SGD), and loss functions (e.g., cross-entropy and MSE losses).
The implemented functions support data splitting into tiles across all possible axes.
The StarPU backend processes tiles according to the underlying task scheduling policies.
Numerical experiments demonstrate that NNTile can train the GPT2 model with $49.9$ billion of parameters on a single node with 8 A100 GPUs. 
In comparison, PyTorch can train a similar model with only $10.6$ billion of parameters on the same node within the same TensorFloat32 precision.
Thus, we confirm the efficiency of task-based parallelism for training large language models with limited computational resources.


\bibliographystyle{unsrt}
\bibliography{lib}


\end{document}